\pdfoutput=1
\documentclass[11pt]{article}

\usepackage[final]{acl}

\usepackage{times}
\usepackage{latexsym}

\usepackage[T1]{fontenc}
\usepackage[utf8]{inputenc}

\usepackage{microtype}

\usepackage{inconsolata}

\usepackage{graphicx}

\usepackage{amsmath}
\usepackage{mathtools}
\usepackage{amsfonts}
\usepackage{multirow}
\usepackage{booktabs}
\usepackage{bbm}
\usepackage{array, makecell}
\usepackage{float}

\title {Exploring the Hidden Capacity of LLMs for One-Step Text Generation}

\author{Gleb Mezentsev \\
  AIRI \\
  Skoltech \\
  \texttt{mezentsev@airi.net} \\\And
  Ivan Oseledets \\
  AIRI \\
  Skoltech \\
  \texttt{oseledets@airi.net} \\}

\begin{document}
\maketitle

\begin{abstract}
A recent study showed that large language\\ models (LLMs) can reconstruct surprisingly long texts -- up to thousands of tokens -- via\\ autoregressive generation from just one trained input embedding. In this work, we explore whether autoregressive decoding is essential for such reconstruction. We show that frozen LLMs can generate hundreds of accurate tokens in just one token-parallel forward pass, when provided with only two learned embeddings. This reveals a surprising and underexplored multi-token generation capability of autoregressive LLMs. We examine these embeddings and characterize the information they encode. We also empirically show that, although these representations are not unique for a given text, they form connected and local regions in embedding space -- suggesting the potential to train a practical encoder. The existence of such representations hints that multi-token generation may be natively accessible in off-the-shelf LLMs via a learned input encoder, eliminating heavy retraining and helping to overcome the fundamental bottleneck of autoregressive decoding while reusing already-trained models.
\end{abstract}

\section{Introduction}

Large language models are typically trained to generate text in an autoregressive manner -- they predict one token at a time based on the previously generated context. Several attempts aim to change this. However, they either require an additional model for candidate generation \citep{leviathan2023fast}, substantial additional fine-tuning of autoregressive LLM \citep{cai2024medusa, stern2018blockwise, gloeckle2024better} or full model retraining \citep{ghazvininejad2019mask, austin2021structured, li2022diffusion}. This leaves an open question -- is it possible to reuse autoregressively pretrained LLM for multi-token generation with minimal to no additional training. We discover a previously undocumented phenomenon, that can help us to achieve this goal.

We found that for any given text of reasonable length, there exists a latent one-vector representation of this text, such that, if a frozen pretrained LLM is conditioned on this representation, it accurately generates the whole text in a single forward pass, without any iterative decoding. In this work, we demonstrate this phenomenon, investigate what those compressed representations encode and whether this finding reveals anything about LLMs' parallel generation capabilities.

\begin{figure}[t]
  \centering
  \setlength{\belowcaptionskip}{-10pt}
  \includegraphics[width=0.45\textwidth]{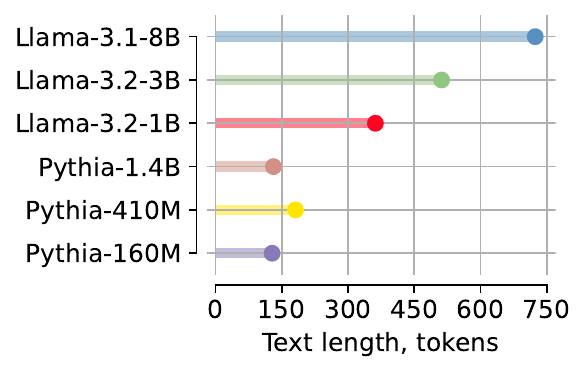}
  \caption{One pass, many tokens. Each dot shows the maximum exact reconstruction length in a single non-autoregressive forward pass with frozen weights, conditioned only on two learned embeddings  -- evidence of hidden multi-token capabilities.}
  \label{fig:best_res}
\end{figure}

Our contribution is as follows:

1. We show that LLMs can reconstruct arbitrary sequences of hundreds of tokens from as few as two learned input embeddings, with one of them being universal for all texts.

2. We identify key design aspects for such a setup, that enable this generation, including the critical importance of input token arrangement.

3. We study how the reconstruction capability varies with the model size and the nature of the target sequence (e.g. natural vs synthetic text).

4. We empirically characterize learned representations -- analyze their information content and embedding-space geometry.

The code is available at \href{https://github.com/Glebzok/OneStepLLMGeneration}{this GitHub page}. 

\section{Related Work}

The most direct influence for our work is a paper by \citet{kuratov2025cramming}, which showed that frozen LLM can reconstruct an arbitrary sequence of tokens $T=[t_1, \dots, t_N]$ if given a sequence of special, so-called memory tokens $[mem_1, \dots, mem_K]$. The embeddings for these tokens are trained by optimizing a causal language modeling objective over a concatenated input sequence ${Z=[mem_1,\dots, mem_K, t_1, \dots, t_N]}$ passed through a frozen LLM. In the case of perfect next-token prediction accuracy (which could be achieved for reasonable text length), this allows the model to autoregressively predict the whole text starting from the memory tokens. The number of memory tokens controls the maximum text length and can be as low as one.

Although surprisingly long (up to 1568 tokens) texts could be compressed even into a single memory token, the authors note that the embeddings trained from different random initializations for the same text end up far apart. Moreover, linear interpolations between those embeddings produce poor reconstruction accuracy, suggesting that the solution space lacks desirable smoothness and locality qualities, which are important for learning a practical encoder that could replace direct optimization.

Our work also relates to efforts in prompt-tuning and its variants \citep{lester-etal-2021-power, liu2024gpt, li-liang-2021-prefix}. Most similarly, \citet{lester-etal-2021-power} train task-specific soft tokens to condition frozen LLMs to improve their performance on new tasks.
% Although, this allows to solve problems the model didn't see during training,
Several speculative \citep{xia-etal-2023-speculative} and parallel \citep{ santilli-etal-2023-accelerating} decoding approaches utilize a similar mechanism for multiple token prediction using decoder architectures. More specifically, they add special [PAD] or [MASK] tokens at the end of the current context in order to make a prediction for several tokens into the future at once. Critically, in these works either special training or multiple generative iterations are required.

Unlike prior work, we show that a frozen LLM can generate accurate multi-token sequences in one forward pass without additional LLM training or iterative decoding.

\section{Method}

To adopt the approach from \citet{kuratov2025cramming} to the non-autoregressive case, we replace all input tokens of the LLM with specially trained "proto-tokens" and predict the target token sequence in one forward pass. In practice, "proto-tokens" are just trainable vectors that are not tied to any real items in the vocabulary. The main difference between regular tokens and these "proto-tokens" is that "proto-tokens" encode multiple tokens at once and only produce human-readable text after passing through the LLM. Our goal is to identify the smallest possible number of such "proto-tokens" needed for accurate reconstruction. Interestingly, we find that it is essential to have at least two -- the performance drops dramatically when using only one (see Section~\ref{seq:results}).

There are many ways to arrange two vectors into an input sequence of arbitrary length. We report results for different variants later in the paper, but here we describe the arrangement that is used in the majority of the experiments.
\paragraph{Exact scheme}

\begin{figure}[h]
  \centering
  \includegraphics[width=0.4\textwidth]{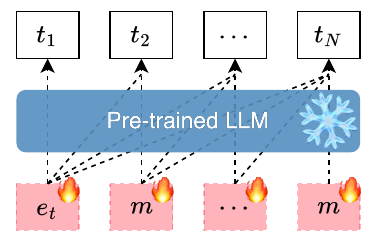}
  \caption{Two "proto-tokens" (trainable embeddings) are fed into frozen, pretrained LLM and optimized in such a way that LLM predicts an arbitrary token-sequence in a single forward pass. $e_t$ is trained for each text separately, while $m$ could be shared across texts.}
  \label{fig:scheme}
\end{figure}

We introduce two "proto-tokens" $e$ and $m$ with trainable embeddings of dimension $d_{model}$ (model input embedding dimension) and construct the input sequence as follows:\\ $Z=[e,m, m, \dots, m]$ -- one copy of token $e$ is followed by $N-1$ copies of token $m$, where $N$ is the target text length. We then train the vectors by optimizing cross-entropy loss between the target sequence $T=[t_1, t_2, \dots, t_N]$ and the frozen LLM's output for the input sequence:
\begin{equation}
    \label{eq:loss}
    L_{CE} = - \sum_{i=1}^N log \mathbb{P}_{LM}(t_i \mid e, \underbracket[0.8pt]{m, \dots, m}_{i-1})
\end{equation}

The prediction is obtained using standard causal attention masking, so that the predicted probabilities for the token $t_i$ depend on the first $i$ input "proto-tokens" (see Figure~\ref{fig:scheme}).

\paragraph{Metrics}
Our main evaluation metric is the number of correctly reconstructed tokens in a generated sequence, defined as:
\begin{equation}
    \label{eq:c_metric}
    C_{tokens} = \sum_{i=1}^N\mathbbm{1}(LM(Z_{[1:i]}) = t_i)
\end{equation}

Additionally, we measure the amount of information contained in the reconstructed token sequence from the perspective of causal language modeling with a given LLM. Specifically, we compute the cross-entropy between the compressed sequence and LLM's autoregressive probability distribution: 
\begin{equation}
    \label{eq:ce_metric}
    H_{LM} = - \sum_{i=1}^{N} log \mathbb{P}_{LM}(t_i \mid t_{<i})
\end{equation}

This quantity measures how uncertain a model is about the compressed text, that is, how much information it contains.

\paragraph{Solution space connectivity}

To gain insights into the structure of the solution space of our problem, we analyze whether different proto-token embeddings obtained for the same text but from different random initializations are connected. We adopt a technique from \citep{garipov2018loss} which is used to find paths connecting different minima of the loss function in computer vision tasks. We optimize the parameters of a degree-two Bezier curve, connecting two solutions, to maximize reconstruction accuracy along the curve. The curve is parameterized by a control point $\pi$ in the following way:
\begin{equation}
    \phi_\pi(\tau) = (1-\tau)^2 p_1 + 2\tau(1-\tau)\pi + \tau^2 p_2
\end{equation}
Here, $p_1$ and $p_2$ are the two original solutions that we aim to connect.

We want to find the value of $\pi$ that minimizes the cross-entropy loss along the curve. To do that, we minimize the expectation of the cross-entropy loss with respect to a uniform distribution of $\tau$:
\begin{equation}
    \label{eq:curve_loss}
    % \begin{aligned}
    l_\pi = \mathop{\mathbb{E}}_{\tau \sim \mathcal{U}[0,1]} {\sum_{i=1}^N -log \mathbb{P}_{LM}(t_i \mid \phi_\pi(\tau))}
    % \end{aligned}
\end{equation}
To do that, we iteratively sample $\tau \sim \mathcal{U}[0,1]$ and optimize $l_\pi$ with respect to $\pi$ using Adam optimizer. This optimization under the uniform distribution over $\tau$ acts as a more tractable alternative to direct optimization under the uniform distribution along the curve itself.

\paragraph{Token sequences similarity}
\label{seq:similarity}
In Section~\ref{sec:interp}, we aim to measure the similarity between two token sequences in order to control for this similarity. To measure token-level similarity we use the cosine distance between TF-IDF embeddings of two sequences. To measure semantic similarity we use cosine-distance between semantic sequence embeddings obtained from a MiniLM model fine-tuned\footnote{\url{https://huggingface.co/sentence-transformers/all-MiniLM-L6-v2}} for the semantic sentence embedding. 

\section{Experiments and results}
\label{seq:results}
We test the ability of different LLMs of varying sizes to generate a predefined text from different sources in a non-autoregressive (parallel) mode. Moreover, we compare different ways to feed our trainable "proto-tokens" into LLM. We also try to understand the structure of the solution space by examining the relations of solutions for different problems.

\paragraph{Models}
We use six models for all experiments: three Pythia \citep{biderman2023pythia} models of sizes 160M, 410M, and 1.4B, and three Llama-3 \citep{grattafiori2024llama} models of sizes 1B, 3B, and 8B.

\paragraph{Data}
Four text sources are used in the experiments to explore the possible connection between reconstruction performance and the text nature.

A set of random texts is generated by sampling from the top 100,000 words of the GloVe vocabulary \citep{pennington-etal-2014-glove}, to evaluate performance on unnatural texts.

To assess generation performance on natural but unseen texts, we use a collection of fanfiction texts from AO3 library \footnote{\url{https://archiveofourown.org/}}, with a publication date cutoff of October 2024, which is later than the end of training for all models. For data processing details, see \citet{kuratov2025cramming}.

The performance on seen natural texts is evaluated using PG-19 dataset \citep{rae2019compressive} -- a part of a dataset used for training Pythia models.

Finally, we include a set of model-specific generated texts. Specifically, for each model and each context text from PG-19 dataset, a suffix of the same length is generated as autoregressive continuation. The generation is done via multinomial sampling with sampling temperature $T=1$.

\paragraph{Training details}
The embeddings of the proto-tokens are initialized from standard normal distribution and optimized via AdamW optimizer \citep{loshchilovdecoupled} with $0.01$ learning rate, $\beta_1$, $\beta_2$ set to $0.9$ and a weight decay of $0.01$. The embeddings are trained for $5000$ iterations with an early stopping at a perfect reconstruction accuracy. This number of iterations is often insufficient for convergence, but due to limited computational resources, we are unable to increase it. Instead, we aggregate results across multiple sequences and report the best results. Although, the exact reconstruction capacity could be under-estimated, we believe that, given he exploratory nature of this work, it is more important to demonstrate and characterize the phenomenon itself, rather than to achieve the precise upper bound on reconstruction capacity. All models are trained using PyTorch framework and Transformers library. Each experimental run is done on a single A100 or H100 80GB GPU with gradient accumulation enabled where necessary. 

\paragraph{Proto-token arrangement}
To select the best way to arrange two proto-tokens as input for LLM for the main experiments, we conduct test runs on a single dataset-model pair for the variety of arrangements. For each arrangement, the same 50 texts from the PG-19 are selected, and the Llama-3.2-1B model is trained on prefixes of these texts at lengths [1, 2, 4, 8, 16, 32, 64, 128, 256, 512, 1024] to assess token-level reconstruction accuracy change with respect to sequence length $N$. A representative selection of results is presented in Table~\ref{tab:schemes}.

\begin{table}[hbp]
\setlength{\tabcolsep}{2.5pt}
\resizebox{\linewidth}{!}{
\begin{tabular}{@{} lllll @{}} 
 \toprule
 Arrangement & $N=1$ & $N=2$ & $N=4$ & $N=256$ \\ 
  \midrule
$\begin{aligned}
[e]_{_{\times N}}\end{aligned}$ & $1.00_{_{\pm 0.00}}$ & $0.45_{_{\pm 0.31}}$ &  $0.17_{_{\pm 0.18}}$ & $0.01_{_{\pm 0.01}}$ \\ 
\midrule
 $\begin{aligned}[e]_{_{\times (N/2)}}[m]_{_{\times  (N/2)}}\end{aligned}$ & $1.00_{_{\pm 0.00}}$ & $1.00_{_{\pm0.00}}$ & $0.12_{_{\pm 0.13}}$ & $0.01_{_{\pm 0.01}}$ \\ 
$\begin{aligned}[e,m]_{_{\times (N/2)}}\end{aligned}$ & $1.00_{_{\pm0.00}}$ & $1.00_{_{\pm0.00}}$ & $1.00_{_{\pm0.00}}$ & $0.17_{_{\pm 0.34}}$ \\ 
 \midrule
 $\begin{aligned}[e][m]_{_{\times N}}\end{aligned}$ & $1.00_{_{\pm0.00}}$ &$1.00_{_{\pm0.00}}$  & $1.00_{_{\pm0.00}}$ & $0.97_{_{\pm 0.15}}$ \\ 
 $\begin{aligned}[e][m]_{_{\times (N-1)}}\end{aligned}$ & $1.00_{_{\pm0.00}}$ & $1.00_{_{\pm0.00}}$ & $1.00_{_{\pm0.00}}$ & $0.99_{_{\pm 0.10}}$ \\ 
 \bottomrule
\end{tabular}
}
\caption{Reconstruction accuracies for different input token arrangements across varying sequence lengths. Subscripts indicate the number of copies for each proto-token. In the second-to-last scheme the LLM is trained to predict the first text token $t_1$ for the proto-token $e$, while with the last one, the prediction for proto-token $e$ is not guided and $t_1$ is a target for the first copy of $m$.}
\label{tab:schemes}
\end{table}

Interestingly, having two proto-tokens is essential. The one-token scheme fails to reconstruct even 2-token text, while best two-token schemes reconstruct $256$-token texts almost perfectly. 

Moreover, the way these two tokens are arranged is also important, with the best results obtained when the first token $e$ is followed by $N-1$ copies of the second token $m$. This asymmetrical arrangement and critical necessity for two tokens suggest possible variation in functions of $e$ and $m$. It is possible, that while one of them mostly incorporates language information, the role of the other one is mainly structural or mechanistic. This could be related to the phenomenon of "attention sinks" -- \citet{xiao2023efficient} showed that LLMs strongly attend to the initial tokens in the sequence even when they are not relevant. So, it is possible, that in order to successfully decode "information" proto-token, LLM needs a distinguishable "sink" proto-token, which can be used as attention sink. 

\paragraph{Token sharing}
\label{seq:token_sharing}
In the previous section, we showed that the quality of reconstruction is very dependent on having two separate proto-tokens as an input. This observation led us to hypothesize that, a second token serves a structural or mechanistic purpose and does not contain information about the sequence itself. In that case, the second token could be shared between texts, reducing the number of optimized parameters, and simplifying the training process of the potential encoder.

To test this hypothesis, we run the same optimization process, splitting $256$ texts from the PG-19 into groups of varying sizes $S_g \in [1, 4, 16, 64, 256]$  and sharing either $e$ or $m$ within each group. We selected the maximum length of the text that can be losslessly compressed in a non-shared mode - $256$. The results are averaged over $10$ random seeds. The selection of results is presented in Table~\ref{tab:sharing}.

\begin{table}[htbp]
\begin{tabular}{@{}lllll@{}}
\toprule
Shared & Agg & $S_g=1$ & $S_g=16$ & $S_g=256$\\
\midrule
\multirow[t]{2}{*}{$e$} & max & $1.00_{_{\pm 0.00}}$ & $0.99_{_{\pm 0.01}}$ & $0.99_{_{\pm 0.02}}$ \\
 & avg & $0.98_{_{\pm 0.08}}$ & $0.90_{_{\pm 0.17}}$ & $0.86_{_{\pm 0.20}}$ \\
\midrule
\multirow[t]{2}{*}{$m$} & max & $1.00_{_{\pm 0.00}}$ & $1.00_{_{\pm 0.00}}$ & $1.00_{_{\pm 0.01}}$ \\
 & avg & $0.98_{_{\pm 0.07}}$ & $0.86_{_{\pm 0.19}}$ & $0.83_{_{\pm 0.18}}$ \\
\bottomrule
\end{tabular}
\caption{Reconstruction accuracy with one of  proto-tokens shared within groups for different group sizes. "max" indicates that for every text, maximum accuracy across ten random seeds is averaged across texts, while "avg" denotes averaging across both seeds and texts.}
\label{tab:sharing}
\end{table}

Sharing either token yields comparable performance if provided with a sufficiently large number of restarts (random seeds), but the required number of restarts  increases significantly with group size.

\begin{table*}[htbp]
\centering
\setlength{\tabcolsep}{2.5pt}
\resizebox{\textwidth}{!}{
\begin{tabular}{@{}lllllllll@{}}
\toprule
& & \multirow{2}{*}{Share $m$} & \multicolumn{3}{c}{Pythia} & \multicolumn{3}{c}{Llama} \\
\cmidrule(lr){4-6} \cmidrule(lr){7-9}
 &  &  & 160M & 410M & 1.4B & 3.2-1B & 3.2-3B & 3.1-8B \\
\midrule
\multirow{4}{*}{Random} & \multirow[c]{2}{*}{$C_{tokens}$} & False & $90$ & $92$ & $90$ & $256$ & $362$ & $512$ \\
 &  & True & $45$ & $22$ & $45$ & $181$ & $256$ & $256$ \\
\cmidrule{2-9}
 & \multirow{2}{*}{$H_{LM}$} & False & $507.5_{_{\pm 105.9}}$ & $377.1_{_{\pm 133.1}}$ & $470.7_{_{\pm 103.1}}$ & $1551.3_{_{\pm 159.5}}$ & $2193.4_{_{\pm 190.2}}$ & $2974.4_{_{\pm 298.3}}$ \\
 &  & True & $247.9_{_{\pm 32.0}}$ & $91.1_{_{\pm 30.8}}$ & $231.0_{_{\pm 37.9}}$ & $947.7_{_{\pm 155.0}}$ & $1292.2_{_{\pm 217.4}}$ & $1309.4_{_{\pm 234.6}}$ \\
\midrule
\multirow{4}{*}{Fanfics} & \multirow[c]{2}{*}{$C_{tokens}$} & False & $128$ & $128$ & $131$ & $362$ & $512$ & $724$ \\
 &  & True & $45$ & $45$ & $45$ & $181$ & $288$ & $362$ \\
\cmidrule{2-9}
 & \multirow{2}{*}{$H_{LM}$} & False & $358.9_{_{\pm 73.3}}$ & $395.4_{_{\pm 97.8}}$ & $261.0_{_{\pm 56.4}}$ & $1107.6_{_{\pm 129.1}}$ & $1408.4_{_{\pm 179.5}}$ & $1763.3_{_{\pm 280.2}}$ \\
 &  & True & $145.0_{_{\pm 26.2}}$ & $82.3_{_{\pm 28.1}}$ & $147.9_{_{\pm 29.7}}$ & $576.4_{_{\pm 90.4}}$ & $835.9_{_{\pm 121.7}}$ & $1112.8_{_{\pm 168.6}}$ \\
\midrule
\multirow{4}{*}{PG-19} & \multirow[c]{2}{*}{$C_{tokens}$} & False & $128$ & $167$ & $128$ & $362$ & $512$ & $724$ \\
 &  & True & $45$ & $32$ & $64$ & $181$ & $256$ & $362$ \\
\cmidrule{2-9}
 & \multirow{2}{*}{$H_{LM}$} & False & $388.4_{_{\pm 66.4}}$ & $408.8_{_{\pm 96.3}}$ & $298.4_{_{\pm 77.4}}$ & $993.8_{_{\pm 183.4}}$ & $1346.0_{_{\pm 218.4}}$ & $1659.8_{_{\pm 344.5}}$ \\
 &  & True & $156.0_{_{\pm 33.9}}$ & $88.1_{_{\pm 30.3}}$ & $156.0_{_{\pm 30.2}}$ & $456.5_{_{\pm 56.5}}$ & $826.1_{_{\pm 117.6}}$ & $832.3_{_{\pm 171.0}}$ \\
\midrule
\multirow{4}{*}{\makecell{PG-19 \\ (gen)}} & \multirow[c]{2}{*}{$C_{tokens}$} & False & $128$ & $181$ & $128$ & $362$ & $512$ & $724$ \\
 &  & True & $45$ & $32$ & $64$ & $181$ & $362$ & $362$ \\
\cmidrule{2-9}
 & \multirow{2}{*}{$H_{LM}$} & False & $354.1_{_{\pm 72.0}}$ & $379.2_{_{\pm 82.6}}$ & $277.6_{_{\pm 71.3}}$ & $927.3_{_{\pm 103.4}}$ & $1266.6_{_{\pm 125.9}}$ & $1653.1_{_{\pm 211.4}}$ \\
 &  & True & $153.0_{_{\pm 17.8}}$ & $106.9_{_{\pm 38.5}}$ & $197.1_{_{\pm 39.3}}$ & $478.7_{_{\pm 85.7}}$ & $788.6_{_{\pm 130.8}}$ & $771.7_{_{\pm 143.0}}$ \\
\bottomrule
\end{tabular}}
\caption{Maximum reconstruction capacities for different models on different datasets.}
\label{tab:main}
\end{table*}

Depending on the proto-token being shared, we can build different intuitions behind the function of the shared tokens and the method itself. 

If the $e$-token is shared, which is located in the very beginning of the input sequence, the analogy that comes to mind is prompt-tuning \cite{lester-etal-2021-power}, where a set of prompt embeddings is trained in order to improve performance in some specific task. In our case, a shared token $e$ could be viewed as an "instruction" saying what an LLM should do with the upcoming embeddings ($m$-tokens) -- decode different pieces of information for different positions.

If the $m$-token is shared, then training and prediction scheme resembles some of the speculative decoding approaches \cite{xia-etal-2023-speculative}, where a number of special [mask] tokens are appended at the end of the sequence and the prediction for all them is then done in parallel.
For all other experiments, unless stated otherwise, we use scheme with sharing $m$ token between texts and random seeds and $e$ token being unique for each text-seed pair.

\paragraph{Generation capacity}
We already see that similar to autoregressive mode \citep{kuratov2025cramming}, LLMs can generate fairly long sequences in just one forward pass. To characterize this capability and understand how it scales with model size and changes depending on the nature of the text, we run the optimization process for text prefixes of the predefined lengths [4, 5, 8, 11, 16, 22, 32, 45, 64, 90, 128, 181, 256, 362, 512, 724, 1024, 1448]. We report the maximum values of $C_{tokens}$ and $H_{LM}$ which correspond to the longest prefix for which at least $0.99$ token-level accuracy is achieved -- we treat such sequences as successfully reconstructed. In addition to a scheme with a shared $m$ token, we also run a scheme with $m$ not shared, to eliminate the effect of the insufficient number of random initializations. While our results in Section~\ref{seq:token_sharing}, suggest that $m$, can in principle, be shared without any quality drop, we also note that the optimization process is highly sensitive to initialization, especially when the proto-tokens are shared. The results are presented in Table~\ref{tab:main} with the best results across datasets presented in Figure~\ref{fig:best_res}.

Larger models in Llama the family show greater reconstruction capabilities than the smaller ones, while the situation with Pythia models is less obvious, with all the models showing approximately the same performance. Llama 1B model is also able to reconstruct almost three times larger sequences compared to Pythia model of the same size.

\begin{figure*}[htbp]
  \includegraphics[width=1\textwidth]{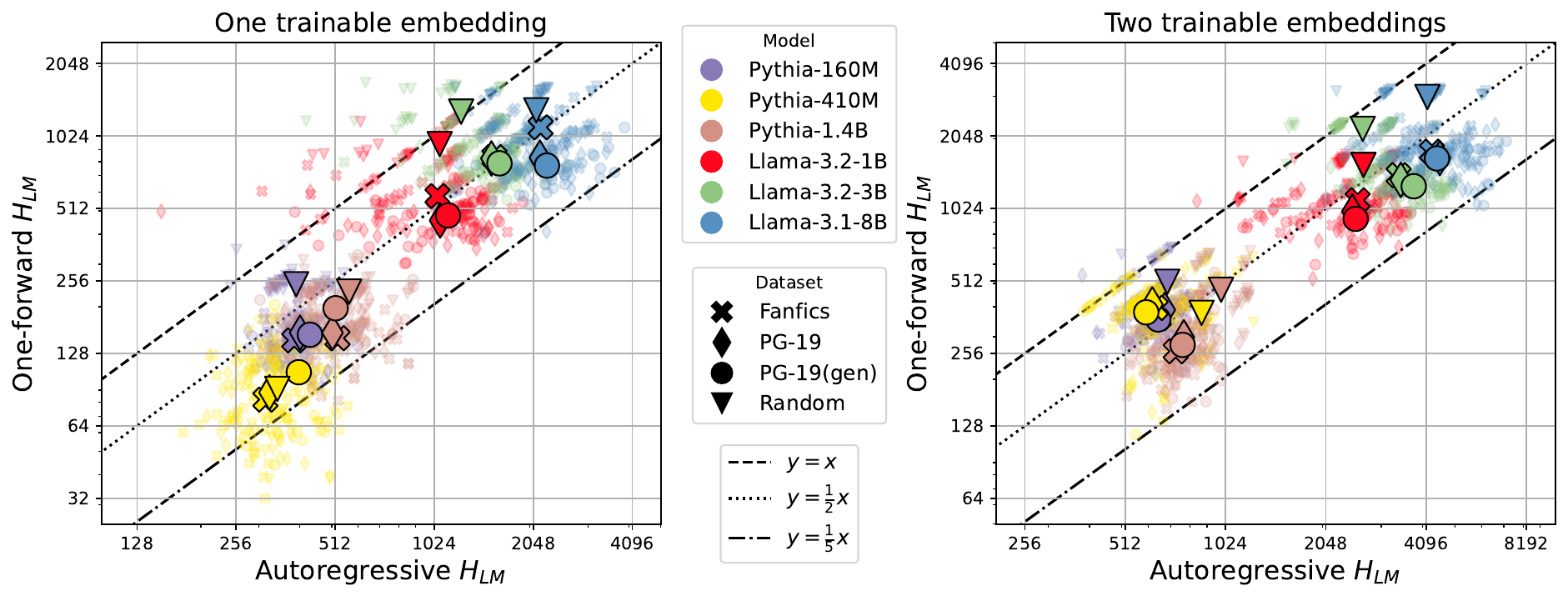}
  \caption{Maximum language information ($H_{LM}$ for a maximum text prefix that is accurately reconstructed) compressed for different models and datasets. In the left plot, a single [mem] token is used in the autoregressive setting, and in the non-autoregressive one, $m$ proto-token is shared between all texts within each model. In the right plot, two [mem] tokens are used and $m$ proto-tokens are not shared. Each small point on the plots represents a single text, larger points indicate the average within each (model, dataset) pair.}
  \label{fig:main}
\end{figure*}

The natural text source (unseen, seen or generated) does not seem to have any systematic influence on the quality of reconstruction in terms of the number of tokens, while for unnatural random texts the generation capacity is significantly worse. This suggests that "proto-tokens" do not "store" tokens directly, but encode some more high-level representations, using language modeling capabilities of LLM. However, we also can not say that the compressibility of the text is determined by its likelihood under the sequential language model. In fact, we observe the opposite trend -- lower total information content $H_{LM}$ is compressed for less information-dense texts, such as generated by the LLM itself. This difference is highlighted in Figure~\ref{fig:main}, where the amount of the information contained in trainable tokens is compared to autoregressive setup. The performance for unnatural texts is very similar and sometimes even identical, while for natural texts, the difference in capacity can be up to five times lower. However, more often the difference is just two-fold, suggesting that autoregressive decoding approximately doubles the "effective" information density in natural text.

\begin{figure}[htbp]
\setlength{\belowcaptionskip}{-10pt}
\includegraphics[width=\linewidth]{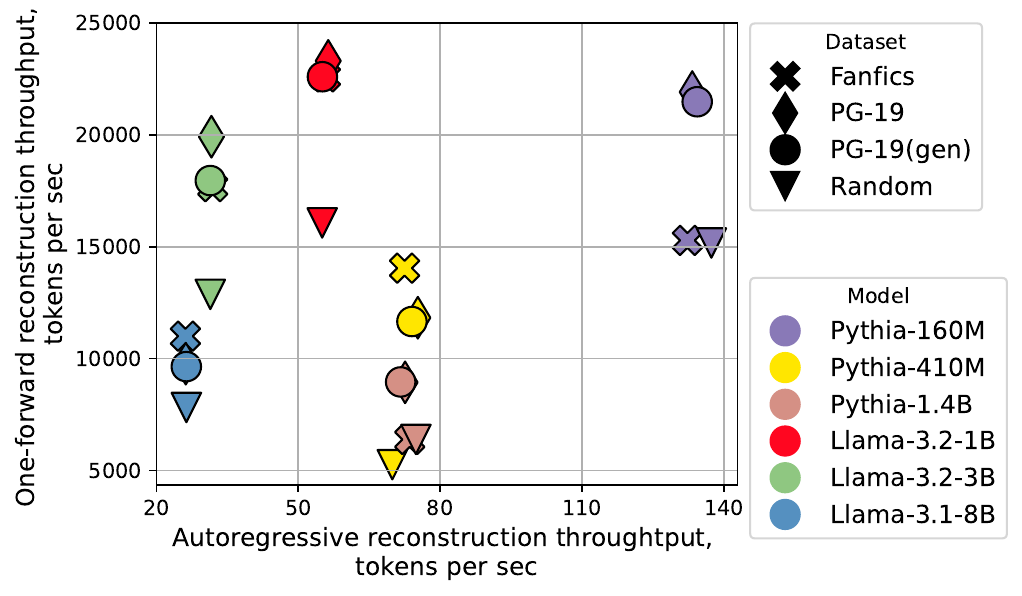}
  \caption{Reconstruction throughput for autoregressive and non-autoregressive setups. For each model-dataset pair, the throughput equals to a maximum losslessly compressible length divided by the reconstruction time.}
  \label{fig:throughtput}
\end{figure}

Although less information-dense, our one-forward method achieves significantly higher decoding throughput in the context of text reconstruction -- outperforming its autoregressive counterpart by a factor of 279 on average (Figure~\ref{fig:throughtput}). This dramatic difference is due to the number of forward passes. While an obvious downstream task is still to be found, such speed could matter in many settings where fast decoding is particularly important.

While we do not introduce the method as a practical way of compressing or generating texts, but rather as a demonstration of interesting phenomenon, we still measure the training time across models and text lengths to demonstrate the full picture. Training time 
(Table~\ref{tab:training_time}) scales roughly linearly with sequence length, with around $10$ seconds for length $32$ and around $200$ seconds for length $512$.

\begin{figure*}[htbp]
\includegraphics[width=\textwidth]{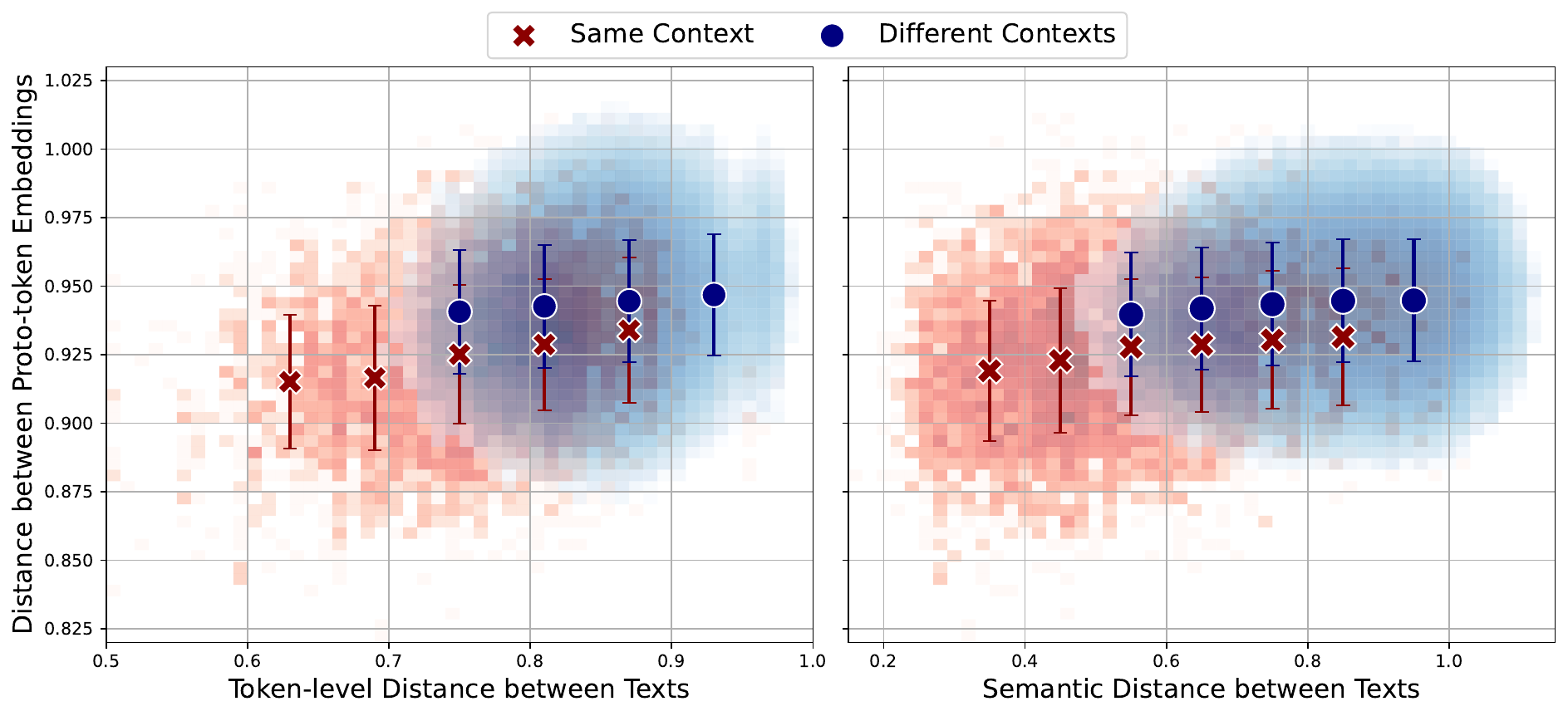}
  \caption{We compare proto-token embedding distances for same context text pairs and different-context text pairs. Token-level distance is measured as cosine distance between TF-IDF embeddings. Semantic distance is measured as cosine distance between semantic text embeddings (see Section~\ref{seq:similarity} for details).}
  \label{fig:suffix_dist}
\end{figure*}

\begin{table}[htbp]
\centering
\small
\setlength{\tabcolsep}{3pt}
\resizebox{0.6\columnwidth}{!}{%
\begin{tabular}{lrrrrr}
\toprule
 Model | N & $32$ & $64$ & $128$ & $256$ & $512$ \\
\midrule
Pythia-160M  & $6$  & $25$ & $68$  & $-$ & $-$ \\
Pythia-410M  & $10$ & $21$ & $106$ & $-$ & $-$ \\
Pythia-1.4B  & $10$ & $66$ & $129$ & $-$ & $-$ \\
Llama-3.2-1B & $11$ & $15$ & $27$  & $87$        & $-$ \\
Llama-3.2-3B & $14$ & $18$ & $28$  & $78$        & $261$ \\
Llama-3.1-8B & $16$ & $20$ & $29$  & $60$        & $215$ \\
\bottomrule
\end{tabular}%
}
\caption{Proto-token training time in seconds for different models and sequence lengths averaged over datasets.}
\label{tab:training_time}
\end{table}

\paragraph{Proto-tokens interpretation}
\label{sec:interp} 

We examine the information encoded in proto-tokens and the implications this has for potential practical applications. In worst case scenario, they directly encode target tokens (imagine a vector containing token\_ids). If so, the entire "language generation" effort happens during encoding, making decoding irrelevant for accelerated inference -- though the approach could still be useful as a context-compression tool. The alternative is that proto-tokens encode a compressed representation of a prefix which, when the model generates from it, produces the observed suffix. In that case, the hard work of text generation is done during decoding, which is more promising from the point of view of accelerated inference. All the intermediate options are also possible.

\begin{figure}[htbp]
\includegraphics[width=\linewidth]{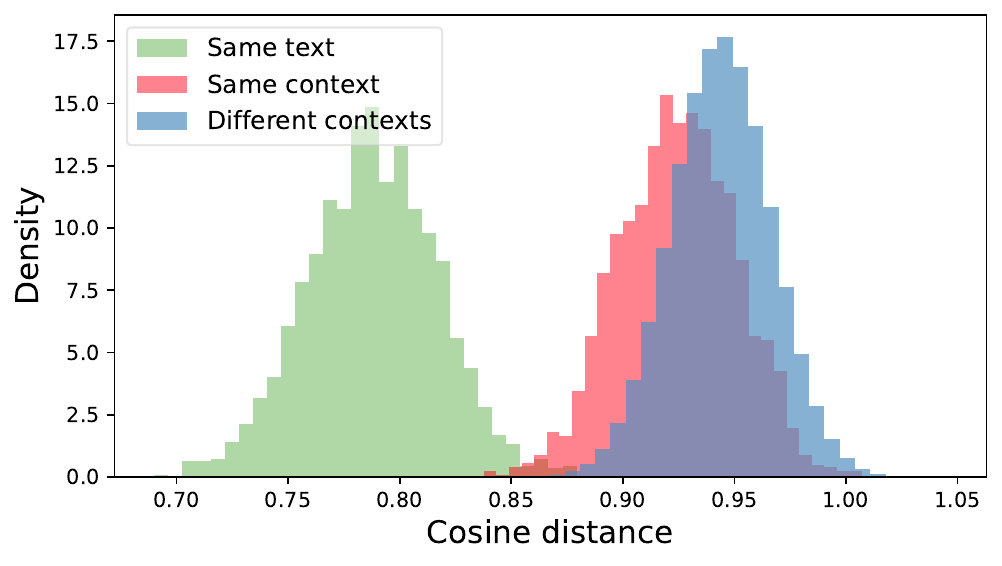}
  \caption{Cosine embedding distances for different pairings of proto-tokens. We select 50 contexts from PG19 and for each context, generate 10 continuation texts. We find one solution for each of the first 9 generations and 10 different-seed solutions for the last generation.}
  \label{fig:angles}
\end{figure}

We start by measuring the distances between three types of proto-token embedding pairs: 1) corresponding to the same generated sequence, but different random seeds, 2) corresponding to the different texts but generated from the same context, 3) corresponding to the different texts generated from different contexts. As shown in Figure~\ref{fig:angles}, the same-text solutions are almost always located closer to each other than different-texts solutions, which suggests locality in the learned representations. At the same time, same-context solutions are noticeably closer to each other than different-context ones. This may indicate that the encoded information at least partially reflects the potential context of the text. However, we should be careful to account for the texts generated from the same context being more similar in general.

To do that, we measure pairwise distances between generated texts, and examine whether the distance between learned proto-token embeddings differ for a fixed distance between the texts. We use token-level measure of text similarity and semantic-level measure (see Section~\ref{seq:similarity}). For both measures, (Figure~\ref{fig:suffix_dist}) we observe that, given the same distances between texts, the proto-token embeddings are on average closer when the texts originate from the same context. We conclude that learned proto-tokens contain information beyond the information about the target sequence itself -- they somehow partially describe the potential context of the sequence. However, we should note that, the effect of having the same context on the distance between proto-tokens is small, and the distributions for same-context distances and different-context distances heavily overlap. Our results suggest that proto-tokens still mostly contain information about the text itself with only a fraction of the information about the context preserved.
% \pagebreak

We also conducted a preliminary experiment on accessing the information contained in proro-tokens without first decoding them into text. We took 50 128-token context sequences from PG-19 dataset, generated 256-token continuations with Llama3.2-1B model and trained $(e,m)$ pairs only for the first 128 tokens of the model continuations. Then we started the autoregressive generation of the same model from different combinations of proto-tokens and \(\langle\text{BOS}\rangle\)-token and visually compared the contents of the resulting token sequences with both contexts and model-continuations. In all cases, the resulting sequences either contain meaningless token-combinations or meaningful texts that are not related to either context or continuation. We conclude that the information from proto-tokens could not be accessed without decoding at least when they are used directly as an autoregressive generation context.

\begin{figure}[h]
\includegraphics[width=\linewidth]{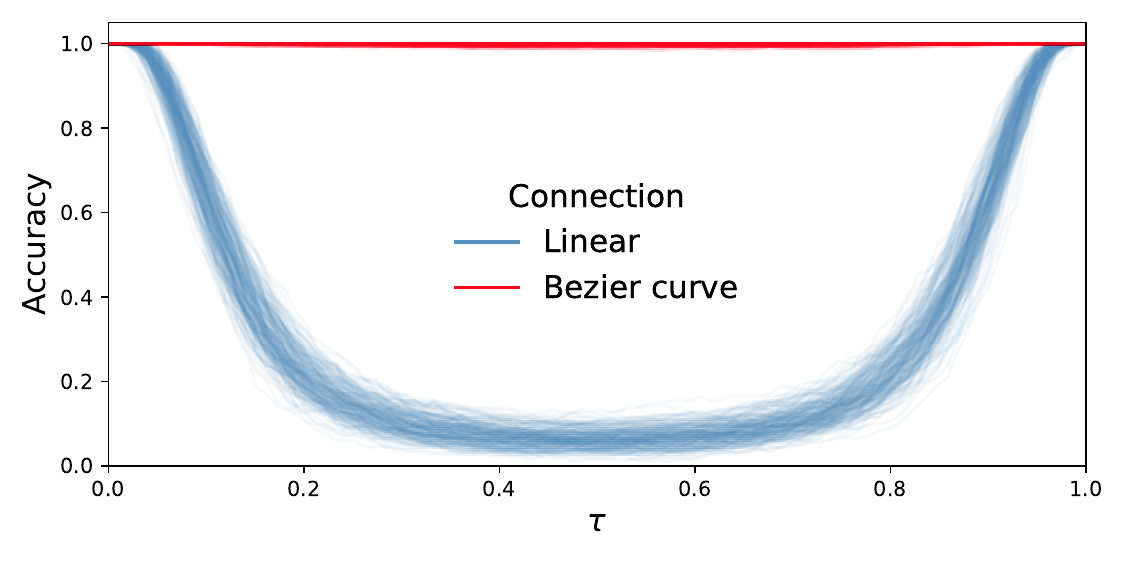}
  \caption{Pairwise interpolation accuracies between $10$ solutions for $5$ texts ($5\times 10 \times 9 / 2$ pairs in total).}
  \label{fig:interpolation}
\end{figure}

\paragraph{Proto-tokens embedding space structure}
\citet{kuratov2025cramming} raised the following concern about the structure of the solution space in the autoregressive setup. Even though the same-text token embeddings are on average closer to each other than different-text token embeddings, they seem to be disconnected -- a linear interpolation between two solutions does not yield a valid reconstruction. This could mean that the potential encoding to this space could be problematic as the same object could be mapped to disconnected regions. We find that in our non-autoregressive case, the linear interpolation between same-text solutions also does not produce a solution (Figure~\ref{fig:interpolation}).

\vspace{-5pt}

However, the solutions could be connected using quadratic Bezier curves (parabolic segments) lying inside "solution set". This means that even though same-text solutions do not form a convex set, they form a connected set. In fact, our experiments show that the maximum ratio between Bezier curve length and the corresponding linear connection is only $1.2$, indicating that the paths are nearly linear. These results demonstrate that the solution space is fairly well behaved, providing reasonable hope that an encoder model could be built to map into that space.

\section{Discussion and Conclusions}
In this paper, we demonstrate that frozen LLMs have a surprising ability to generate hundreds of accurate tokens in a single forward pass -- without any iterative decoding -- when provided with just two specially trained "proto-tokens".

We find that both the number and the arrangement of such tokens is crucial for enabling this generation capacity. Interestingly, with only one proto-token, LLMs are unable to generate more than a single token of text. In contrast, two properly arranged proto-tokens can enable the generation of sequences hundreds of tokens long. This significant leap in the performance, along the observation that one of the vectors can (in principle) be shared across many texts, suggests that proto-tokens play different functional roles during generation. 
% However, the precise nature of the role differentiation remains unclear.

We find that bigger model size does not universally imply better generation capacity. While larger models in Llama-3 family demonstrate improved reconstruction capacity, Pythia models show no such trend -- larger models do not perform better.
% Whether this difference is connected to the architectural variations is an open question. 

Additionally, we do not observe any consistent relationship between the source of the natural text and the reconstruction ability of LLMs. Surprisingly, even for the texts generated by the LLM itself, the number of successfully reconstructed tokens is the same as for any other natural text. However, with random-token sequences, performance drops noticeably. This suggests that our reconstruction process does not fully leverage the language modeling capabilities of LLMs, and may instead mostly rely on low-level token patterns.
% \newpage

Although the reconstructed sequences in the non-autoregressive setting are, on average, two times shorter than those in the autoregressive case, the efficiency of single-forward approach allows to achieve up to 279× greater generation throughput.

We also observe that proto-tokens encode more than just the target sequence. Embeddings of the "proto-tokens" corresponding to the different texts generated from the same context are, on average, closer to each other than those from unrelated sequences. This indicates that learned representations capture some potential contextual information.

Finally, we discover that the embedding space of proto-tokens has very desirable structural properties -- proto-tokens corresponding to the same text, form localized and connected regions with smooth transitions via quadratic interpolation. These findings suggest that it may be feasible to build an encoder capable of mapping into this space, opening the door to future work on non-autoregressive inference and representation learning.

We view this work as an existence proof: certain text representations can elicit multi-token behavior in frozen, single-token LLMs. Making them practical requires training an encoder that maps text into these representations. As a future work, we plan to study decoders that either generate a suffix from an encoded prefix or reconstruct the entire text. Depending on the setup, such systems could enable multi-token or chunk-wise generation, learned compression or RAG (potentially by extracting information directly without decoding).

\section{Limitations}\indent

1. Lack of immediate practical application: Most importantly, this work highlights an interesting quirk of LLMs and does not suggest any immediate practical implications or real-life usages for the method yet, as direct proto-token optimization should be replaced with parametrized encoder for any practical application.

2. Architectural dependence: The method demonstrates different behavior across model families, suggesting some architectural dependence. As a result, our method may potentially not generalize to other model architectures.

3. Limited domain coverage: While we evaluate four different text sources, the results may not generalize beyond those explored in our experiments.

4. Evidence, not bounds: Our 5k-step optimization budget may not always reach full convergence. Thus, our results should be read as evidence of feasibility (existence of one-pass decoding) rather than a precise capacity bound.
% \section{Broader Impact}

\section*{Acknowledgments}
The work was supported by the grant for research centers in the field of AI provided by the Ministry of Economic Development of the Russian Federation in accordance with the agreement 000000C313925P4F0002 and the agreement with Skoltech №139-10-2025-033.

\bibliography{anthology,custom}

% \appendix

% \section{Example Appendix}
% \label{sec:appendix}

% This is an appendix.

\end{document}